\documentclass[letterpaper]{article} 
\usepackage{aaai24}  
\usepackage{times}  
\usepackage{helvet}  
\usepackage{courier}  
\usepackage[hyphens]{url}  
\usepackage{graphicx} 
\usepackage{natbib}  
\usepackage{caption} 
\usepackage{algorithm}
\usepackage{algorithmic}

\usepackage{amsmath, amsfonts}

\usepackage[caption=false,font=normalsize,labelfont=sf,textfont=sf]{subfig}
\usepackage{multirow}
\usepackage{booktabs}

\usepackage{newfloat}
\usepackage{listings}
\DeclareCaptionStyle{ruled}{labelfont=normalfont,labelsep=colon,strut=off} 
\floatstyle{ruled}
\newfloat{listing}{tb}{lst}{}
\floatname{listing}{Listing}
\nocopyright
\title{Weakly-Supervised Temporal Action Localization by\\ Inferring Salient Snippet-Feature }
\author{
 Wulian Yun,
    Mengshi Qi,
    Chuanming Wang,
    Huadong Ma
}
\affiliations{
    Beijing Key Laboratory of Intelligent Telecommunications Software and Multimedia, \\Beijing University of Posts and Telecommunications, China\\
\{yunwl,qms,wcm,mhd\}@bupt.edu.cn
}

\usepackage{bibentry}

\begin{document}

\maketitle

\begin{abstract}
Weakly-supervised temporal action localization aims to locate action regions and identify action categories in untrimmed videos simultaneously by taking only video-level labels as the supervision. Pseudo label generation is a promising strategy to solve the challenging problem, but the current methods ignore the natural temporal structure of the video that can provide rich information to assist such a generation process. In this paper, we propose a novel weakly-supervised temporal action localization method by inferring salient snippet-feature. First, we design a saliency inference module that exploits the variation relationship between temporal neighbor snippets to discover salient snippet-features, which can reflect the significant dynamic change in the video. Secondly, we introduce a boundary refinement module that enhances salient snippet-features through the information interaction unit. Then, a discrimination enhancement module is introduced to enhance the discriminative nature of snippet-features. Finally, we adopt the refined snippet-features to produce high-fidelity pseudo labels, which could be used to supervise the training of the action localization network. Extensive experiments on two publicly available datasets, \emph{i.e.}, THUMOS14 and ActivityNet v1.3, demonstrate our proposed method achieves significant improvements compared to the state-of-the-art methods. Our source code is available at https://github.com/wuli55555/ISSF.
\end{abstract}

\section{Introduction}
\label{sec:intro}

Temporal action localization (TAL)~\cite{7780488,8237579,8578222,Huang_2022_CVPR,he2022asm} aims to find action instances from untrimmed videos, \emph{i.e.}, predicting the start positions, end positions, and categories of certain actions. It is an important yet challenging task in video understanding and has been widely used in surveillance and video summarization. 
To achieve accurate localization, most existing methods~\cite{7780488,8237579,8578222,Lin_2018_ECCV,8953536} rely on training a model in a fully supervised manner with the help of human-labeled precise temporal annotations. However, fine-detailed labeling of videos is labor-intensive and expensive. In contrast, weakly-supervised methods recently have gained increasing attention from both academia and industry, since they only utilize video-level labels for temporal action localization, achieving competitive results while reducing the cost of manual annotations.

\begin{figure}[t]
\begin{center}
\setlength{\fboxrule}{0pt}
\setlength{\fboxsep}{0cm}

\fbox{\rule{0pt}{0in} \rule{.0\linewidth}{0pt}
\hspace{-4mm}
     \includegraphics[width=1.1\linewidth]{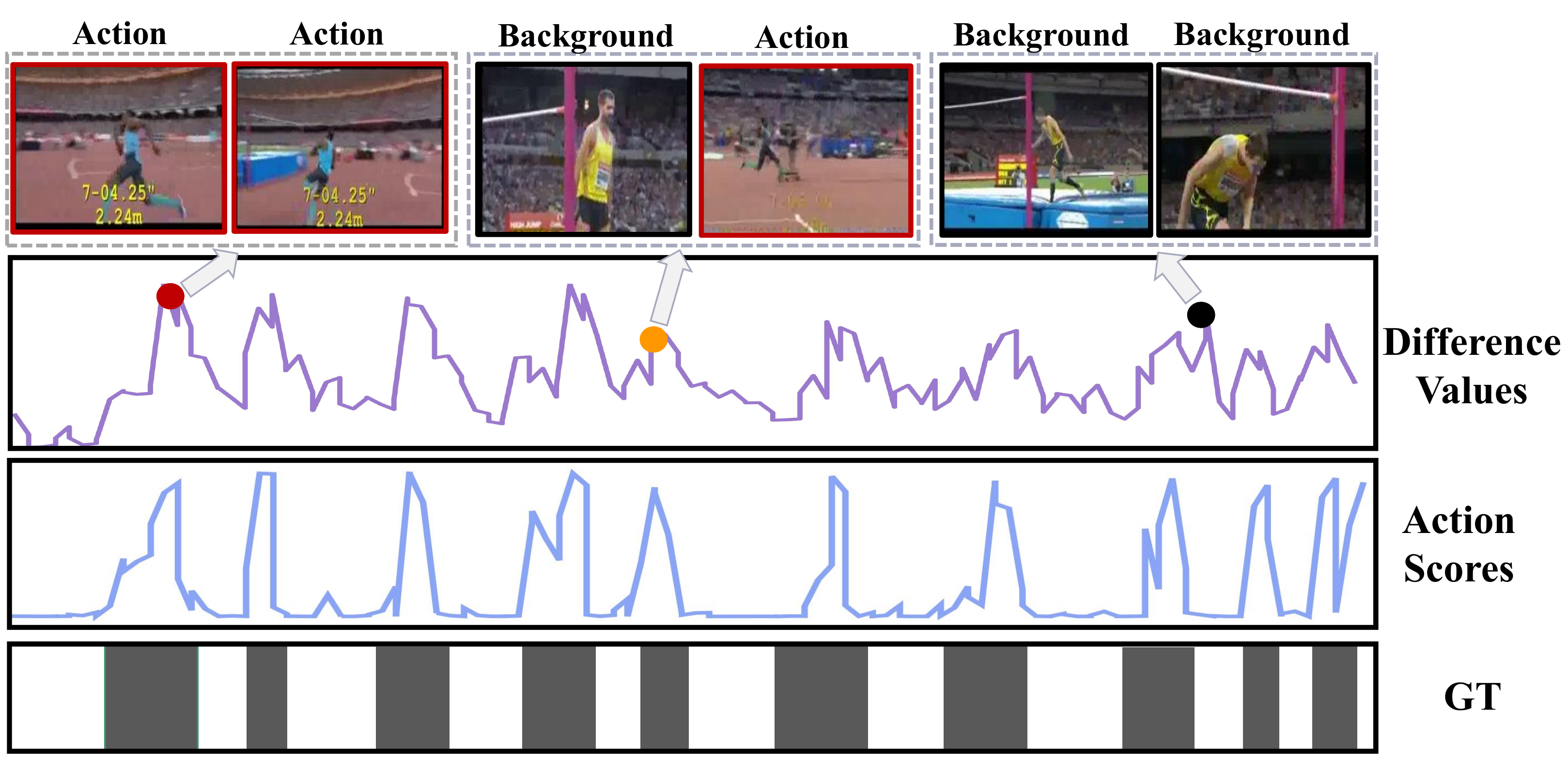}}
\end{center}
\vspace{-2mm}
    \caption{Illustration of difference values among snippets, action scores and Ground-Truth~(GT). The action and background snippets are marked as red and black boxes, respectively.}
    \vspace{-3mm}
\label{fig:1}
\end{figure}

Weakly-supervised TAL methods
~\cite{10.1609/aaai.v33i01.33019070,Shi_2020_CVPR,qu_2021_acmnet,DBLP:conf/aaai/LeeWLB21,9578220,DBLP:conf/iccv/NarayanCHK0021} 
mainly utilize a ``localization by classification'' framework, where a series of Temporal Class Activation Maps~(TCAMs)~\cite{8578804,paul2018w} are obtained by snippet-wise classification, and then TCAMs are used to generate temporal proposals for action localization. 
However, the classifiers primarily tend to focus on easily distinguishable snippets while ignoring other subtle yet equally important information, so there is a discrepancy between classification and localization.
To balance the performance of classification and localization,
pseudo label based methods
~\cite{Huang_2022_CVPR, he2022asm,9423165,10.1007/978-3-030-58539-6_3,DBLP:conf/eccv/LuoGSKWDX20,Li_2022_CVPR} 
have been proposed, which supervises the training of the model mainly by generating snippet-level pseudo label information.

Nevertheless, accurately generating pseudo label remains challenging, since existing methods ignore the important role played in the temporal structure of videos. 
We observed that neighbor snippets exhibit obvious distinctively difference relationships, which can discover salient features and identify differentiated boundaries. As shown in Figure~\ref{fig:1}, neighbour snippet-features with substantial variations~(higher difference value) may correspond to the junctions between action and background, alternations between action, or abrupt changes between background.
However, how to find these features and refine them into more discriminative features is the key to discover action boundaries and then improve the localization performance.

Inspired by this observation, we propose a novel weakly-supervised TAL method, which takes a new perspective that boosts the generation of high-fidelity pseudo-labels by leveraging the temporal variation. 
First, we design a saliency inference module to discover significant snippet-feature by leveraging the variation and calculating the difference values of neighbor snippet pairs.
However, this process only considers local relationships and ignores the global information in the video.
Thus, we propose a boundary refinement module to enhance salient features through information interaction while making the model focus on the entire temporal structure.
Subsequently, considering diverse action information can provide additional clues, we propose a discrimination enhancement module to further refine the feature by constructing a memory to introduce the same category of action knowledge.
Finally, the output features are fed into the classification head to generate the final refined pseudo labels for supervision.

The contributions can be summarized as follows:

\par\textbf{(1)} We propose a new pseudo-label generation strategy for weakly-supervised TAL by inferring salient snippet-feature, which can exploit the dynamic variation.

\par\textbf{(2)} We design a boundary refinement module and a discrimination enhancement module to enhance the discriminative nature of action and background, respectively.

\par\textbf{(3)} We conduct extensive experiments and the results show our model achieves 46.8 and 25.8 average mAP on THUMOS14 and ActivityNet v1.3, respectively.

\section{Related Work}
\noindent
\textbf{Fully-supervised temporal action localization.}
Fully-supervised TAL has been an active research area in video understanding ~\cite{9351755,9052709,8954105,8621027,liu2016large,liu2018t} for many years and existing methods are divided into two categories,~\emph{i.e.,} one-stage methods and two-stage methods. One-stage methods
~\cite{8953536,10.1145/3123266.3123343,9171561,Lin_2021_CVPR}
predict action boundaries as well as labels simultaneously. On the contrary, two-stage methods
~\cite{7780488,8237579,8578222,9009541} first find candidate action proposals and then predict their labels. However, these fully supervised methods are trained with instance-level human annotation, leading to an expensive and time-consuming process.

\noindent
\textbf{Weakly-supervised temporal action localization.}
Weakly-supervised TAL methods~\cite{10.1609/aaai.v33i01.33019070,DBLP:conf/eccv/MinC20,Shi_2020_CVPR,DBLP:conf/aaai/LeeWLB21,9578220,DBLP:conf/iccv/NarayanCHK0021,10.1007/978-3-030-58539-6_3, Huang_2022_CVPR, mengyuan2022ECCV_DELU} 
mainly learn from video-level labels, which avoid labor-intensive annotations compared to the fully-supervised methods. UntrimmedNet~\cite{inproceedingsU} and STPN~\cite{8578804} generate class activation sequences by Multiple Instance Learning (MIL) framework and then locate action instances by thresholding processing. RPN ~\cite{DBLP:conf/aaai/HuangHOW20a} and 3C-Net~\cite{9008791} use metric learning algorithms to learn more discriminative features. Lee~\emph{et al.}~\cite{lee2020BaS-Net} design a background suppression network to suppress background snippets activation. 
However, there is still a discrepancy between classiﬁcation and localization. Recently, numerous methods~\cite{9423165,DBLP:conf/eccv/LuoGSKWDX20,10.1007/978-3-030-58539-6_3,9577332, Huang_2022_CVPR,he2022asm} attempt to generate pseudo labels to supervise the model and thus alleviate the discrepancy.
RefineLoc~\cite{9423165} alleviates the discrepancy between classification and localization by extending the previous detection results to generate pseudo labels.
Luo~\emph{et al.}~\cite{DBLP:conf/eccv/LuoGSKWDX20} exploit the Expectation–Maximization framework~\cite{543975} to generate pseudo labels by alternately updating the key-instance assignment branch and the foreground classification branch.
TSCN~\cite{10.1007/978-3-030-58539-6_3} generates frame-level pseudo labels by later fusing attention sequences in consideration of two-stream consensus.
Li~\emph{et al.}~\cite{Li_2022_CVPR} exploit contrastive representation learning to enhance the feature discrimination ability.
ASM-Loc~\cite{he2022asm} generates action proposals as pseudo labels by using the standard MIL-based methods.
In contrast, our method exploits the variation between neighbor snippet-features to find salient snippet-features, and further designs a boundary refinement module and a discrimination enhancement module to generate high-fidelity pseudo labels.

\begin{figure*}[t]
\begin{center}
\setlength{\fboxrule}{0pt}
\setlength{\fboxsep}{0cm}
\vspace{-4mm}
\hspace{-2.5mm}
     \includegraphics[width=0.99\linewidth]{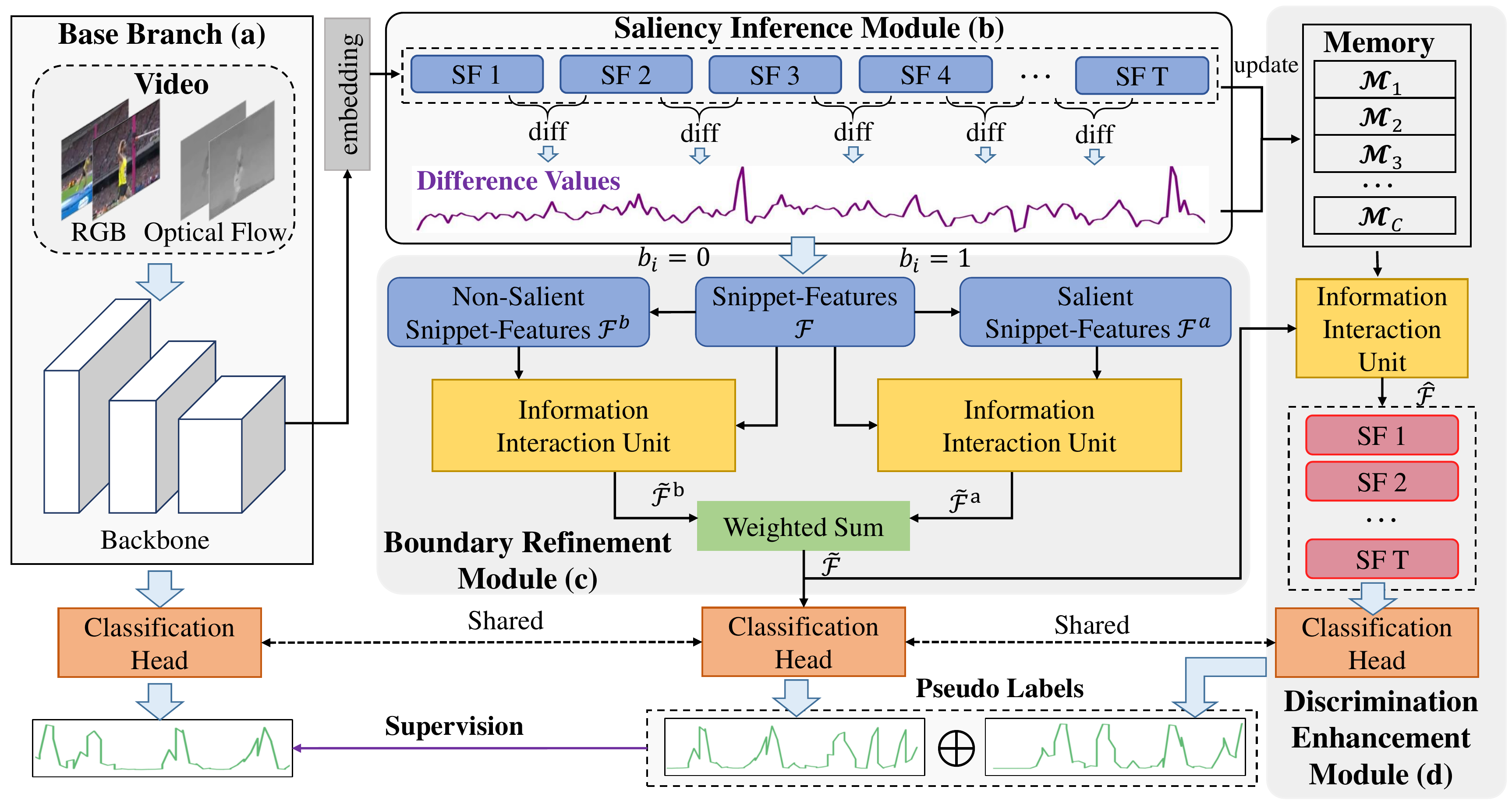}
\end{center}
\vspace{-3mm}
\caption{Overview of our model. Firstly, the base branch (a) extracts features from RGB and optical flow in a video and uses the classification head to predict TCAMs. Then, the saliency inference module~(b) exploits the variation relationship between snippet-features to discover salient snippet features. Next, the boundary refinement module (c) utilizes the information interaction unit to enhance salient snippet features. Subsequently, the discrimination enhancement module (d) leverages action information stored in memory to enhance the discrimination of action and background. Finally, (c) and (d) generate high-fidelity pseudo labels to supervise the base branch.}
\vspace{-3mm}
\label{fig:2}
\end{figure*}


\section{Methodology}
In this section, we will begin by presenting the problem definition of weakly-supervised TAL and provide an overview of our proposed method. Next, we will describe the different modules of our method in detail, which are designed to generate high-fidelity pseudo labels by utilizing the variation between snippet-features. Finally, we introduce the training details of optimizing the temporal localization model.

\noindent
\textbf{Problem definition.}
Weakly-supervised TAL aims to predict a group of action instances ($c$, $q$, $t_s$, $t_e$) for each test video with the assistance of a set of untrimmed training videos $\{V_i\}^N_{i=1}$ and their corresponding ground-truth labels $\{{y_i}\}^N_{i=1}$. Specifically, $y_i \in \mathbb{R}^C$ is a binary vector indicating the presence/absence of each of $C$ actions. For one action instance, $c$ denotes the action category, $q$ refers to the prediction confidence score, $t_s$ and $t_e$ mean the start time and end time of the action, respectively.

\noindent
\textbf{Overview.}
The overview of our proposed method is shown in Figure~\ref{fig:2}, which mainly contains four parts: \emph{(a) base branch}, \emph{(b) saliency inference module},  \emph{(c) boundary refinement module}, and \emph{(d) discrimination enhancement module}.

First, in the base branch, we exploit a fixed pre-trained backbone network~(e.g., I3D) to extract $T$ snippet-features from both the appearance (RGB) and motion (optical flow) of the input video. Then, a learnable classification head is adopted to classify each snippet and obtain the predicted TCAMs.
Second, we utilize the saliency inference module to generate salient snippet-features by calculating the difference between adjacent pairs of snippet-features. 
Subsequently, the boundary refinement module and the discrimination enhancement module both utilize the information interaction unit to refine coarse boundaries by enhancing salient snippet-features and the separability of action snippet-features from those of the background. 
Finally, the output features are fed into the classification head to generate high-fidelity pseudo labels as a supervised signal for the base branch.

\subsection{Base Branch}
Given an untrimmed video $V$, we follow~\cite{8578804,Huang_2022_CVPR} to split it into multiple non-overlapping snippets $\{v_i\}^T_{i=1}$, and then we use the I3D~\cite{8099985} network pre-trained on the Kinetics-400~\cite{DBLP:journals/corr/KayCSZHVVGBNSZ17} dataset to extract features from the RGB and optical flow streams for each snippet.
An embedding layer takes the concatenation of these two types of features to fuse them together, and the fused features of all snippets are treated as snippet-features of the video $\mathcal{F}=\{f_1, f_2, \cdots, f_T\} \in \mathbb{R}^{T \times D}$, where T is the number of snippets and D denotes the dimension of one snippet-feature.

Next, we use the classification head to obtain Temporal Class Activation Maps~(TCAMs) $\mathcal{T} \in \mathbb{R}^{T \times (C+1)}$, where $C+1$ denotes the number of action categories plus the background class. 
Specifically, following previous work~\cite{Huang_2022_CVPR}, the classification head consists of a Class-agnostic Attention~(CA) head and a Multiple Instance Learning~(MIL) head.

\subsection{Saliency Inference Module}

The significant variation of temporal neighbor snippets can indicate whether each snippet belongs to a salient snippet-feature. Therefore, we propose a saliency inference module that utilizes such variation to explore the difference between neighbor snippet pairs and then use it to identify salient boundaries in the video.

Given a video and its snippet-level representation $\mathcal{F} \in \mathbb{R}^{T \times D}$, we first calculate the difference value $\tau_{(t-1,t)}$ of each pair of temporal adjacent snippet-features $\{f_{t-1}$, $f_{t}\}$ in the formulation of: 
\begin{equation}
\label{eq:2}
\tau_{(t-1,t)}= \sum_{d=1}^{D}|\mathrm{diff}(f_{t}, f_{t-1}, d)|,
\end{equation}
where $\mathrm{diff}$ denotes the operation of dimensional-wise subtraction, and $d\in D$ means the element index of the feature. 
Subsequently, we obtain the difference set $\tau$ of the input video by calculating the difference for all pairs:
\begin{equation}
\tau=\{\tau_{(1,2)},\tau_{(2,3)},\cdots,\tau_{(t-1,t)}\}.
\end{equation}

To obtain the salient snippet-features of the video, we first perform a descending sort on the difference set $\tau$, and then assign the initial labels $\mathcal{B}=\{b_{i}\}_{i=1}^{T}$ to each snippet based on the sorted $\tau$. The snippets with the top $K$ sorted scores are selected as salient snippet-features, while the remaining are selected as non-salient snippet-features, and the process of assigning labels can be formulated as:
\begin{equation}
    b_{t} = \left\{ 
        \begin{aligned} 
        1&,\ \text{if} \ \tau_{(t-1,t)}\in \text{Top}({\operatorname{sorted}}(\tau), K) \\
        0&,\ otherwise \\
        \end{aligned} 
    \right
    .,
\end{equation}
where $b_t=1$ denotes that its corresponding snippet $f_t$ belongs to the salient snippet-features, otherwise to the non-salient snippet-features. 
Finally, salient snippet-features are discovered in a simple manner. However, since these snippet-features cannot be determined as actions or backgrounds, directly using these features to supervise the learning of the base branch may lead to poor performance. Next, we will present how to refine these salient snippet-features.

\subsection{Boundary Refinement Module}
In the saliency inference module, we calculate the difference values between each pair of adjacent snippets, and the operation can be seen as one type of exploiting local relationships, but the relationship among non-local snippets is still underexplored. Therefore, we propose a boundary refinement module to enhance salient snippet-features, where exploring the contextual relationship among the salient snippet-features, non-salient snippet-features, and the same video snippet-features via information interaction unit along the channel and temporal dimensions, respectively.

First, we collect the salient snippet-features~($b_i$=1) and non-salient snippet- features~($b_i$=0) candidates to form $\mathcal{F}^a\in\mathbb{R}^{T^a\times D}$ and $\mathcal{F}^b\in\mathbb{R}^{T^b\times D}$, respectively, where $\mathcal{F}^a \cup \mathcal{F}^b = \mathcal{F}$ , $T^a + T^b = T$, $T^a$ denotes the number of salient snippet-features, and $T^b$ denotes the number of non-salient snippet-features.
Then, we leverage a channel-wise information interaction unit in the squeeze-and-excitation pattern
to generate the feature $\hat{\mathcal{F}}^{a} \in \mathbb{R}^{T^a \times D}$:
\begin{equation}
\label{eq:5}
\hat{\mathcal{F}}^{a}= \frac{\exp \left(\theta(\mathcal{F}^a)\right)}{\sum_{d=1}^D \exp\left(\theta(\mathcal{F}^a_{\cdot, d})\right)} \otimes \mathcal{F}^{a} + \mathcal{F}^a,
\end{equation}
where $\otimes$ denotes the element-wise multiplication. $\theta$ is a simple multi-layer perceptron, which is consisted of \textit{FC-ReLU-FC}. We set the weight of the first \textit{FC} to ${\mathbf{W}_1} \in \mathbb{R}^{D \times (D/r)}$ and that of the second \textit{FC} to $ {\mathbf{W}_2} \in \mathbb{R}^{(D/r) \times D}$, and $r$ is a scaling factor.
Residual connection is adopted to maintain the stability of training. 

Subsequently, we conduct a temporal-level information interaction unit to capture the global contextual relationships between $\hat{\mathcal{F}}^{a}$ and $\mathcal{F}$ 
as the following equation:
\begin{equation}
\label{eq:6}
\tilde{\mathcal{F}}^a= \operatorname{softmax} (\mathcal{F} \odot (\hat{\mathcal{F}}^a)^T) \odot \hat{\mathcal{F}}^a,
\end{equation}
where $\odot$ denotes the matrix multiplication. 
By integrating such information, we obtain a set of discriminative snippet-features $\mathcal{\tilde{F}}^a \in \mathbb{R}^{T\times D}$. 

However, some information contained in $\mathcal{F}^b$ maybe neglected, which contains some action-related or background-related information. Thus, utilizing the information in $\mathcal{F}^b$ can help boost the diversity of snippet-features, and we also utilize the temporal-level information interaction unit to generate non-salient enhanced features $\tilde{\mathcal{F}}^{b}$  between $\mathcal{F}^{b}$ and $\mathcal{F}$ through Eq.(\ref{eq:5}) and Eq.(\ref{eq:6}). Note that the parameters in Eq.\eqref{eq:5} are not shared between $\mathcal{F}^a$ and $\mathcal{F}^b$.

Finally, we apply a weighted sum operation to balance the contribution between $\tilde{\mathcal{F}}^{a}$ and $\tilde{\mathcal{F}}^{b}$ to obtain the enhanced features $\tilde{\mathcal{F}} \in \mathbb{R}^{T\times D}$ as follows:
\begin{equation}
\tilde{\mathcal{F}}=\operatorname{sum}(\tilde{\mathcal{F}}^{a}, \tilde{\mathcal{F}}^{b}, \sigma) = \sigma  \tilde{\mathcal{F}}^{a}+(1-\sigma)  \tilde{\mathcal{F}}^{b},
\end{equation}
where $\sigma$ denotes a trade-off factor. 

\subsection{Discrimination Enhancement Module}
Action information from videos of the same category can provide additional clues to help improve the discriminative nature of the snippet-features and the quality of the generated pseudo-labels. Therefore, we design a discrimination enhancement module that utilizes the correlation among videos to make action and background snippet-features more separable.

First, we introduce a memory bank $\mathcal{M} \in \mathbb{R}^{C \times N \times D}$ as the action knowledge base to store the diverse action information from the entire dataset during training, where $C$ denotes the number of classes, $N$ indicates the number of stored snippets of each class, and D is the dimension number.
Initially, we use the classification head to predict the scores of the salient snippet-features and select the snippets with the highest $N$ classification scores to initialize the memory $\mathcal{M}$ along with the scores. At $t$-th training iteration, we select $N$ snippet-features $\mathcal{F}^{(t)}_{[c]}$ with the high scores for each class to update the memory of last iteration $\mathcal{M}_{[c]}^{(t-1)}$. The process can be formulated as:

\begin{equation}
\mathcal{M}_{[c]}^{(t)} \leftarrow(1-\eta) \cdot \mathcal{M}_{[c]}^{(t-1)}+\eta \cdot \mathcal{F}_{[c]}^{(t)},
\end{equation}
where $\eta$ denotes the momentum coefficient. To boost the robustness, we exploit the momentum update strategy~\cite{9157636} to update memory $\mathcal{M}$, so $\eta$ is adjusted by: 
\begin{equation}
\eta=\eta_0 \cdot \log \left(\exp \left({e}/{E}\right)+1\right),
\end{equation}
where $\eta_0$ denotes the initial momentum coefficient, $e$ is the current epoch, $E$ denotes the total epoch,
and $c$ is the class index of the current snippet. Meanwhile, we use the temporal-level information interaction units to implement the interaction between the mixed features $\tilde{\mathcal{F}}$ in the boundary refinement module and memory $\mathcal{M}_{[c]}^{(t)}$ to bring the class information of the entire dataset into $\tilde{\mathcal{F}}$, which can be formulated as:
\begin{equation}
\hat{\mathcal{F}} =\operatorname{softmax} (\tilde{\mathcal{F}} \odot (\mathcal{M}_{[c]}^{(t)})^{T}) \odot \mathcal{M}_{[c]}^{(t)}.
\end{equation}

Finally, we get the output features $\tilde{\mathcal{F}}$ and $\hat{\mathcal{F}}$ from the boundary refinement module and discrimination enhancement module. Then, we feed them to the classification head to output two TCAMs, \emph{i.e.,} $\tilde{\mathcal{T}}$ and $\hat{\mathcal{T}}$, which are summed after to obtain ${\mathcal{T}^p}$ as the pseudo labels to supervise the learning of the base branch.

\begin{table*}[ht]
\centering
\renewcommand{\arraystretch}{1.1}
\setlength{\tabcolsep}{0.7mm}{
\begin{tabular}{c|l|c|ccccccc|ccc}
\specialrule{0.1em}{1pt}{1pt}
\hline  
\multirow{2}{*}{Sup}  & \multirow{2}{*}{~Method} & \multirow{2}{*}{Feature} & \multicolumn{7}{c|}{mAP@IoU(\%)}               & AVG       & AVG       & AVG       \\ \cline{4-13} 
                             &                         &                              & 0.1  & 0.2  & 0.3  & 0.4  & 0.5  & 0.6  & 0.7  & (0.1:0.5) & (0.3:0.7) & (0.1:0.7) \\ \hline
\multirow{5}{*}{Full}        & S-CNN \cite{7780488}                  & -                    & 47.7 & 43.5 & 36.3 & 28.7 & 19.0 & 10.3 & 5.3  & 35.0      &   19.9        & 27.3      \\
                             & SSN \cite{8237579}                    & -                    & 66.0 & 59.4 & 51.9 & 41.0 & 29.8 & -    & -    & 49.6      & -         & -         \\
                             & TAL-Net \cite{8578222}                & -                    & 59.8 & 57.1 & 53.2 & 48.5 & 42.8 & 33.8 & 20.8 & 52.3       & 39.8      & 45.1         \\
                             & GTAN \cite{8953536}                   & -                    & 69.1 & 63.7 & 57.8 & 47.2 & 38.8 & -    & -    & 55.3      & -         & -         \\ \hline 
\multirow{2}{*}{Weak*}       & STAR \cite{10.1609/aaai.v33i01.33019070}                   & I3D                   & 68.8 & 60.0 & 48.7 & 34.7 & 23.0 & -    & -    & 47.0      & -         & -         \\
                             & 3C-Net \cite{9008791}                 & I3D                    & 59.1 & 53.5 & 44.2 & 34.1 & 26.6 & -    & 8.1  & 43.5      & -         & -         \\ 
                             \hline
\multirow{18}{*}{Weak}       
& STPN~\cite{8578804}                    & I3D                    & 52.0 & 44.7 & 35.5 & 25.8 & 16.9 & 9.9  & 4.3  & 35.0      & 18.5     & 27.0      \\
                             & RPN \cite{DBLP:conf/aaai/HuangHOW20a}                    & I3D                    & 62.3 & 57.0 & 48.2 & 37.2 & 27.9 & 16.7 & 8.1  & 46.5      & 27.6      & 36.8      \\
                             & BaS-Net \cite{lee2020BaS-Net}                 & I3D                    & 58.2 & 52.3 & 44.6 & 36.0 & 27.0 & 18.6 & 10.4 & 43.6      & 27.3     & 35.3      \\
                             & DGAM  \cite{Shi_2020_CVPR}                  & I3D                    & 60.0 & 56.0 & 46.6 & 37.5 & 26.8 & 17.6 & 9.0  & 45.6      & 27.5      & 37.0      \\
                             & TSCN \cite{10.1007/978-3-030-58539-6_3}                   & I3D                   & 63.4 & 57.6 & 47.8 & 37.7 & 28.7 & 19.4 & 10.2 & 47.0      & 28.8      & 37.8      \\
                             & A2CL-PT  \cite{DBLP:conf/eccv/MinC20}               & I3D                    & 61.2 & 56.1 & 48.1 & 39.0 & 30.1 & 19.2 & 10.6 & 46.9      & 29.4      & 37.8      \\
                             & UM  \cite{DBLP:conf/aaai/LeeWLB21}                    & I3D                   & 67.5 & 61.2 & 52.3 & 43.4 & 33.7 & 22.9 & 12.1 & 51.6      & 32.9      & 41.9      \\
                             & CoLA~\cite{zhang2021cola}                   & I3D                    & 66.2 & 59.5 & 51.5 & 41.9 & 32.2 & 22.0 & 13.1 & 50.3      & 32.1      & 40.9      \\
                             & AUMN \cite{AUMN}                   & I3D                    & 66.2 & 61.9 & 54.9 & 44.4 & 33.3 & 20.5 & 9.0  & 52.1      & 32.4      & 41.5      \\
                              & UGCT \cite{9577332}                   & I3D                    & 69.2 & 62.9 & 55.5 & 46.5 & 35.9 & 23.8 & 11.4  & 54.0      & 34.6      & 43.6      \\
                             & D2-Net  \cite{DBLP:conf/iccv/NarayanCHK0021}                & I3D                    & 65.7 & 60.2 & 52.3 & 43.4 & 36.0 & -    & -    & 51.5      & -         & -         \\
                             & FAC-Net \cite{Huang2021ForegroundActionCN}                & I3D                    & 67.6 & 62.1 & 52.6 & 44.3 & 33.4 & 22.5 & 12.7 & 52.0      & 33.1      & 42.2      \\
                             & DCC~\cite{Li_2022_CVPR}                     & I3D                    & 69.0 & 63.8 & 55.9 & 45.9 & 35.7 & 24.3 & 13.7 & 54.1      & 35.1      & 44.0      \\
                             & RSKP~\cite{Huang_2022_CVPR}                    & I3D                    & 71.3 & 65.3 & 55.8 & 47.5 & 38.2 & 25.4 & 12.5 & 55.6      & 35.9      & 45.1      \\
                             & ASM-Loc ~\cite{he2022asm}                & I3D                    & 71.2 & 65.5 & 57.1 & 46.8 & 36.6 & 25.2 & 13.4 & 55.4      & 35.8      & 45.1      \\ 
                              & DELU~\cite{mengyuan2022ECCV_DELU} & I3D & 71.5 & 66.2 & 56.5 & 47.7 & 40.5 & \textbf{27.2} & \textbf{15.3} & 56.5      & 37.4      & 46.4 \\
                              & A-TSCN~\cite{9826380}   & I3D  & 65.3  & 59.0 & 52.1 & 42.5 & 33.6 & 23.4 & 12.7  & 50.5 & 32.9  & 41.2 \\
                            &  FBA-Net ~\cite{10115434}                & I3D                    & 71.9 & 65.8 & 56.7 & 48.6 & 39.3 & 26.4 & 14.2 & 56.5       &  37.0      & 46.1      \\ 
                              \cline{2-13}   
                             & \textbf{Ours}                    & I3D                            &  \textbf{72.4}    &  \textbf{66.9}    &   \textbf{58.4}     &  \textbf{49.7}   &  \textbf{41.8}    & 25.5     &  12.8    &  \textbf{57.8}  &  \textbf{37.6}       & \textbf{46.8}     \\\hline   
                                                    \specialrule{0.1em}{1pt}{1pt}
\end{tabular}
\caption{Comparison with state-of-the-art methods on THUMOS14 dataset. The AVG columns show average mAP under IoU thresholds of 0.1:0.5, 0.3:0.7 and 0.1:0.7. I3D denotes the utilization of the I3D network as the feature extractor, respectively.  * indicates the methods use extra information. The best results are highlighted in bold. Sup means supervision manner.} 
\label{tab:1}  
}
\end{table*}

\subsection{Training loss}

Following previous methods, the whole learning process is jointly driven by video-level classification loss $\mathcal{L}_\mathrm{cls}$, 
knowledge distillation loss $\mathcal{L}_\mathrm{kd}$ and attention normalization loss $\mathcal{L}_\mathrm{att}$~\cite{10.1007/978-3-030-58539-6_3}. 
The total loss function can be formulated as:
\begin{equation}
\mathcal{L}=\mathcal{L}_\mathrm{c l s}
+ \mathcal{L}_\mathrm{k d}+\lambda\mathcal{L}_\mathrm{a t t},
\end{equation}
where $\lambda$ denotes trade-off factors. 
The knowledge distillation $\mathcal{L}_\mathrm{kd}$ in~\cite{Huang_2022_CVPR} is used to implement the process of ${\mathcal{T}^p}$ supervising $\mathcal{T}$ for training.
The video-level classification loss is the combination of two losses calculated from the CA head and MIL head, which can be formulated as:
\begin{equation}
\mathcal{L}_\mathrm{cls}=\mathcal{L}_\mathrm{CA} + \theta\mathcal{L}_\mathrm{MIL},
\end{equation}
\noindent where $\theta$ is a hyper-parameter.
More details about each loss function please refer to the corresponding references.

\section{Experiments}
\subsection{Datasets and Evaluation Metrics. }
We conduct our experiments on the two commonly-used benchmark datasets, including THUMOS14~\cite{THU} and AcitivityNet v1.3~\cite{7298698}. Following the general weak-supervised setting, we only use the video-level category labels in the training process.

\textbf{THUMOS14} includes 200 untrimmed validation videos and 212 untrimmed test videos, where videos are collected from 20 action categories. Following the previous work ~\cite{inproceedingsU,he2022asm,Huang2021ForegroundActionCN}, we use the validation videos to train our model and test videos for evaluation.

\textbf{ActivityNet v1.3} contains 10,024 training videos, 4,926 validation videos, and 5,044 testing videos of 200 action categories.
Following~\cite{DBLP:conf/aaai/LeeWLB21,Huang_2022_CVPR}, we use the training videos to train our model and validation videos for evaluation.

\textbf{Evaluation metrics.} 
We evaluate the performance of our method with the standard evaluation metrics: mean average precise (mAP) under different intersection over union (IoU) thresholds. For THUMOS14 dataset, we report the mAP under thresholds IoU=\{0.1, 0.2, 0.3, 0.4, 0.5, 0.6, 0.7\}. For ActivityNet v1.3 dataset, we report the mAP under thresholds [0.5:0.05:0.95]. At the same time, we also calculate the average mAP for different IoU ranges on the two datasets.
 
\subsection{Implementation Details}
We implement our model with the PyTorch framework and train the model with Adam optimizer~\cite{2014Adam}. The scaling factor $r$ is set to 4. The hyper-parameter $\theta$ and $\lambda$ are set to 0.2 and 0.1, respectively. The feature is extracted using the I3D~\cite{8099985}, which is pre-trained on the Kinetics-400~\cite{DBLP:journals/corr/KayCSZHVVGBNSZ17} dataset.
For THUMOS14 dataset, we train 180 epochs with a learning rate of 0.00005, the batch size is set to 10, $\sigma$ is set to 0.88, and $K$ is set to $\lfloor50\%*T\rfloor$, where T is the number of video snippets.
For ActivityNet v1.3 dataset, we train 100 epochs with a learning rate of 0.0001, the batch size is set to 32, $\sigma$ is set to 0.9, and $K$ is set to $\lfloor90\%*T\rfloor$.

\subsection{Comparison with State-of-the-Art Methods}
\noindent
\textbf{THUMOS14}. We first compare our method with the state-of-the-art (SOTA) methods on THUMOS14 dataset. These SOTA methods contain fully-supervised methods
and weakly-supervised methods,
the results are shown in Table~\ref{tab:1}. We can observe that our proposed model outperforms the SOTA weakly-supervised temporal action localization methods.
Our proposed method reaches 57.8 at average mAP for IoU thresholds 0.1:0.5, 37.6 at average mAP for IoU thresholds 0.3:0.7, and 46.8 at average mAP for IoU thresholds 0.1:0.7. 
Meanwhile, our result can reach 41.8 at mAP@0.5.
The reasons for the improved performance stem from 1) our method uses the variation relationships between snippet-features to generate salient snippet-features and then considers contextual information to enhance salient snippet-features,  thereby improving the discriminative of snippet-features; 2) we introduce additional clues to fully leverage the relationships between videos, improving the discriminative nature of the action and background snippet-features. Thus, generating more high-fidelity pseudo labels can significantly improve the performance of action localization.

\noindent
\textbf{ActivityNet v1.3}. Table~\ref{tab:2} shows the evaluation results in terms of mAP@IoU on ActivityNet v1.3 dataset. From the table, our model achieves competitive performance compared to other SOTA methods. In addition, our method achieves 25.8 for average mAP, which is 0.7 higher than ASM-Loc, demonstrating the superiority of our method.

\begin{table}[ht]
\renewcommand{\arraystretch}{1.0}
\setlength{\tabcolsep}{0.6mm}{
\centering
\begin{tabular}{l|cccc}
\specialrule{0.1em}{1pt}{1pt}
\hline
\multirow{2}{*}{Method} & \multicolumn{4}{c}{mAP@IoU(\%)} \\ \cline{2-5} 
                                                  & 0.5    & 0.75   & 0.95  & AVG   \\ \hline
STPN~\cite{8578804}                                    & 29.3   & 16.9   & 2.6   & 16.3  \\
CMCS~\cite{8953341}                               & 34.0   & 20.9   & 5.7   & 21.2  \\
BaS-Net~\cite{lee2020BaS-Net}                                & 34.5   & 22.5   & 4.9   & 22.2  \\
TSCN~\cite{10.1007/978-3-030-58539-6_3}                                & 35.3   & 21.4   & 5.3   & 21.7  \\
A2CL-PT~\cite{DBLP:conf/eccv/MinC20}                                 & 36.8   & 22.0   & 5.2   & 22.5  \\
TS-PAC~\cite{9578220}                              & 37.4   & 23.5   & 5.9   & 23.7  \\
UGCT~\cite{9577332}                               & 39.1   & 22.4   & 5.8   & 23.8  \\
AUMN~\cite{AUMN}                                 & 38.3   & 23.5   & 5.2   & 23.5  \\
FAC-Net~\cite{Huang2021ForegroundActionCN}                                  & 37.6   & 24.2   & 6.0   & 24.0  \\
DCC~\cite{Li_2022_CVPR}                                      & 38.8   & 24.2   & 5.7   & 24.3  \\
RSKP~\cite{Huang_2022_CVPR}                                     & 40.6   & 24.6   & 5.9   & 25.0  \\
ASM-Loc~\cite{he2022asm}                                    & \textbf{41.0}   & 24.9   & 6.2   & 25.1  \\ 
A-TSCN~\cite{9826380}  & 37.9 & 23.1 &5.6 & 23.6  \\
\hline
Ours                                         & 39.4       & \textbf{25.8}       & \textbf{6.4}      &  \textbf{25.8}    \\ \hline
\specialrule{0.1em}{1pt}{1pt}
\end{tabular}
\caption{Comparison with state-of-the-art methods on ActivityNet v1.3 dataset. The AVG column shows the averaged mAP under the IoU thresholds [0.5:0.05:0.95].}
\label{tab:2}
}
\vspace{-0.3em}
\end{table}

\subsection{Ablation Study}
We conduct ablation studies to demonstrate the impact of different components in our method on THUMOS14 dataset. 

\noindent
\textbf{Impact of Saliency Inference Module.}
To find a proper function in Eq.\eqref{eq:2}, we explore several strategies to calculate the difference between each pair of neighbor snippets, including cosine distance, $\text{L}_1$ distance, and $\text{L}_2$ distance, and the results are reported in Table~\ref{tab:3}. 
In addition, we explore other ways of generating salient snippet-features, such as random assignment and classification. Among them, random assignment means randomly assigning salient or non-salient labels to each snippet, and classification uses the pre-trained classification head of the base model to classify snippets into salient and non-salient.
The results show that $\text{L}_2$ distance can achieve higher mAP than cosine distance, and $\text{L}_1$ distance yields the best results compared to other methods, so we adopt it as the default \textit{diff} function. 
The reason is that $\text{L}_1$ focuses on the subtle variations between features by computing the absolute differences, which is important for TAL. Whereas cosine distance calculates relative differences and $\text{L}_2$ may suppress these subtle differences by squaring and then taking the square root.

\begin{table}[t]
\centering
\renewcommand{\arraystretch}{1.0}
\setlength{\tabcolsep}{2.2mm}{
        \begin{tabular}{l|ccccc}
        \specialrule{0.1em}{1pt}{1pt}
        \hline
 \multirow{2}{*}{Method} & \multicolumn{5}{c}{mAP@IoU($\%$)} \\ \cline{2-6} 
                                & 0.1           & 0.3           & 0.5           & 0.7           & AVG           \\ \hline
            random       & 68.6          & 53.1          & 34.4          & 11.7          & 42.3          \\    
        classification       & 67.7          & 53.4          & 36.9          & 11.7          & 43.0          \\                        
        cosine distance       & 68.6          & 53.5          & 36.6          & 12.3          & 43.3          \\
        $L_2$ distance          & 70.7          & 55.5          & 36.6          & 12.3          & 44.2          \\
        $L_1$ distance~(Ours)    & \textbf{72.4} & \textbf{58.4} & \textbf{41.8} & \textbf{12.8} & \textbf{46.8} \\ \hline
        \specialrule{0.1em}{1pt}{1pt}
        \end{tabular}
    }
    \caption{Ablation studies about different strategies of detecting salient snippet-feature on THUMOS14 dataset. }
        \label{tab:3}
\vspace{-0.5em}
\end{table}

\begin{table}[t]
\centering
\renewcommand{\arraystretch}{1.0}
\setlength{\tabcolsep}{1.95mm}{
    \begin{tabular}{l|ccccc}
    \specialrule{0.1em}{1pt}{1pt}
    \hline
    \multirow{2}{*}{Method} & \multicolumn{5}{c}{mAP@IoU($\%$)} \\ \cline{2-6} 
                                    & 0.1           & 0.3           & 0.5           & 0.7           & AVG           \\ \hline
    Base       & 62.7          & 45.5          & 29.3          & 10.4          & 37.1          \\
    Base + BRM          & 66.3          & 49.7          & 32.6         & 11.3          & 40.5          \\
    Base + BRM + DEM       & \textbf{72.4} & \textbf{58.4} & \textbf{41.8} & \textbf{12.8} & \textbf{46.8} \\ 
    \hline
    \specialrule{0.1em}{1pt}{1pt}
    \end{tabular}
    \caption{The effects of different modules on THUMOS14 dataset. BRM and DEM denote boundary refinement module and discrimination enhancement module, respectively.}
    \label{tab:4}
}
\vspace{-0.3em}
\end{table}

\begin{table}[t]
\centering
\renewcommand{\arraystretch}{1.0}
\setlength{\tabcolsep}{1.986mm}{
    \begin{tabular}{l|ccccc}
    \specialrule{0.1em}{1pt}{1pt}
    \hline
    \multirow{2}{*}{Method} & \multicolumn{5}{c}{mAP@IoU(\%)} \\ \cline{2-6} 
                            & 0.1    & 0.3    & 0.5   & 0.7   & AVG   \\ \hline
   
    self                    & 65.8   & 48.1   & 30.6   & 10.9  &  39.2      \\
    w/o salient                    & 49.9   & 34.8   & 20.3  & 5.9  &  27.6  \\
    w/o non-salient                    & 71.7   & 56.9   & 39.9   & 12.4  & 45.9       \\
    salient + non-salient                & 71.0   & 54.1   & 35.0   & 10.3  &  43.1       \\
     
    weighted sum   &  \textbf{72.4}     &   \textbf{58.4}        &  \textbf{41.8}       &  \textbf{12.8}         & \textbf{46.8}       \\ 
    temporal-level  & 71.4  & 57.4   & 39.5   & 12.8 & 46.0 \\
    \hline
    \specialrule{0.1em}{1pt}{1pt}
    \end{tabular}
    \caption{The effect of different components in boundary refinement module on THUMOS14 dataset.}
        \label{tab:5}
    }  
\vspace{-4mm}
\end{table}
\begin{table}[t]
\centering
\renewcommand{\arraystretch}{1.0}
\setlength{\tabcolsep}{2.mm}{
    \begin{tabular}{l|ccccc}
    \specialrule{0.1em}{1pt}{1pt}
    \hline
    \multirow{2}{*}{Method} & \multicolumn{5}{c}{mAP@IoU(\%)} \\ \cline{2-6} 
                                                    & 0.1 & 0.3   & 0.5   & 0.7  & AVG   \\ \hline
    direct update
    & 71.9  & 58.1   & 40.3   & 12.3 & 46.4 \\
    momentum update                            & 71.0   & 55.7   & 37.4  & 11.3  & 44.5\\\hline
    \textbf{Ours}   &  \textbf{72.4}     &   \textbf{58.4}        &  \textbf{41.8}       &  \textbf{12.8}         & \textbf{46.8}       \\ \hline
    \specialrule{0.1em}{1pt}{1pt}
    \end{tabular}
    \caption{The effect of different memory update strategies on THUMOS14 dataset.} 
    \label{tab:6}
}
\end{table}

\noindent
\textbf{Impact of Different Modules.}
We evaluate the effect of the boundary refinement module and discrimination enhancement module. The results are presented in Table~\ref{tab:4}. We set the base branch as \textit{Base} and progressively add the boundary refinement module and discrimination enhancement module to Base, and the performances are continuously improved by 3.4 and 6.3 for average mAP.

\noindent
\textbf{Impact of Boundary Refinement Module.}
We evaluate the impact of different variants of  boundary refinement module. The results are reported in Table~\ref{tab:5}, in which
1) \textit{self} denotes utilize temporal-level information interaction unit to interact information between video snippet-features $\mathcal{F}$ and itself;
2) \textit{w/o salient} and \textit{w/o non-salient} denote removing the salient and non-salient snippet-features, respectively;
3) \textit{salient + non-salient} denotes directly adding the two types of features together; 
4) \textit{weighted sum} denotes using weighted sum operation to fuse two types of features; 
5) \textit{temporal} denotes enhancing snippet-features only using the temporal-level information interaction unit.
We can observe that 1) the salient snippet-features have a significant influence on the performance, removing them will lead to a significant drop in the performance; 
2) weighted sum is more effective compared to directly adding, which can assist the utilizing of the information in non-salient snippet-features;
3) information interaction unit at both the channel-level and temporal-level can enhance the discriminative nature of features better.

\noindent
\textbf{Impact of Memory Update Strategies.} We explore the impact of different memory update strategies in the discrimination enhancement module. 
The evaluation results are shown in Table~\ref{tab:6}. We evaluate two variants of memory update strategy, \emph{i.e.}, only using the high-confidence action snippet-features to direct update memory, and only using the momentum update strategy.
From the table, we can see that our method obtains better performance than only using the momentum update strategy, because the momentum update strategy will include many noisy features and impair the learning of intra-video relation. 
The results indicate that our method effectively incorporates more action information compared to the direct update strategy.

\begin{figure}[htbp]
\vspace{-2.5mm}
\begin{center}
    \setlength{\fboxrule}{0pt}
    \setlength{\fboxsep}{0cm}
    \fbox{\rule{0pt}{0in} \rule{.0\linewidth}{0pt}
    \includegraphics[width=1\linewidth, height=0.27\textwidth]{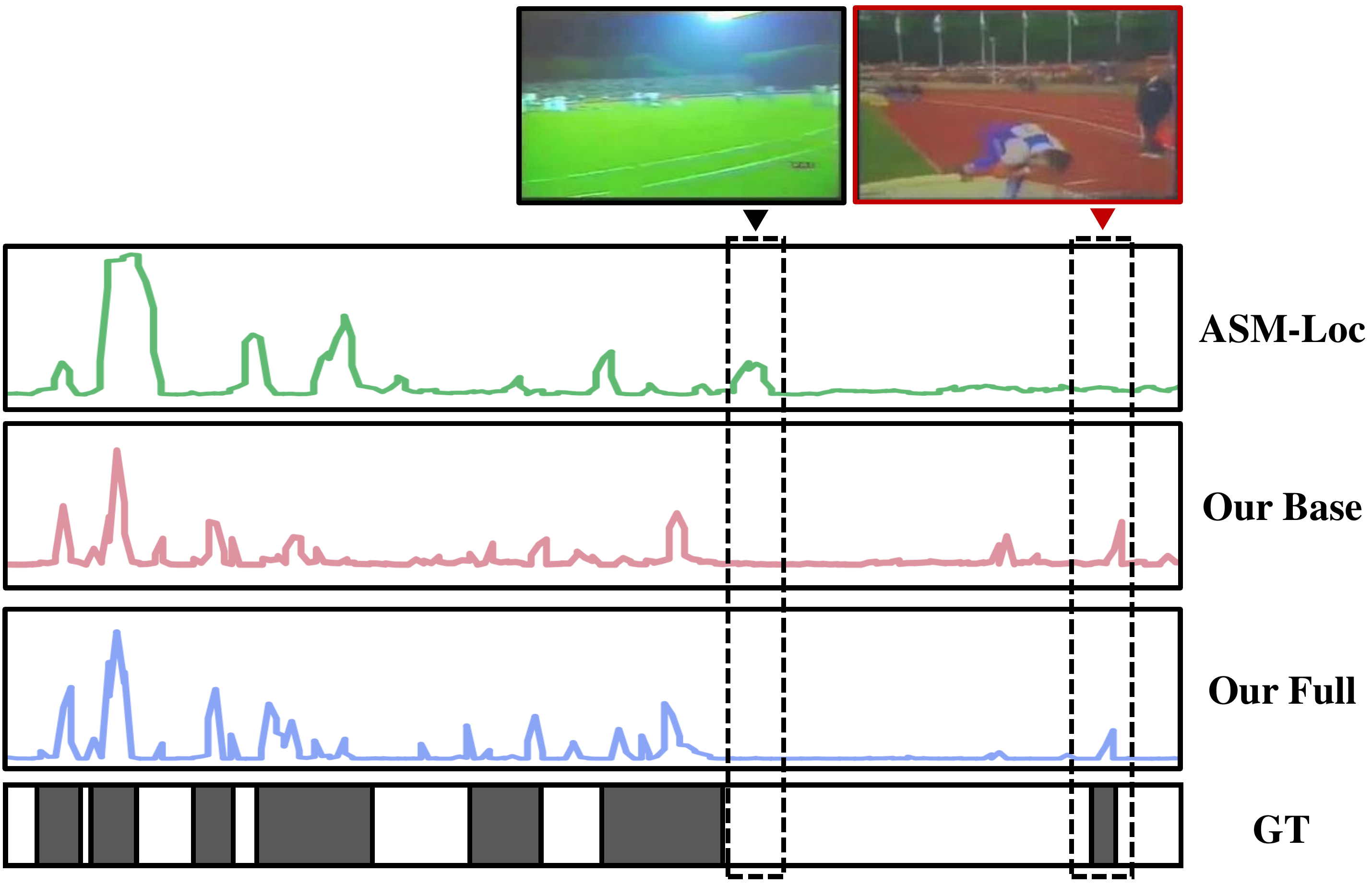}}
\end{center}
\vspace{-4mm}
\caption{Qualitative comparisons of our method, our Base~(base branch), and ASM-Loc on “Shotput" on THUMOS14.} 
\label{fig:3}
\end{figure}

\begin{figure}[t]
\begin{center}
\setlength{\fboxrule}{0pt}
\setlength{\fboxsep}{0cm}
\fbox{\rule{0pt}{0in} \rule{.0\linewidth}{0pt}
\hspace{-3.0mm}
    \includegraphics[width=0.98\linewidth, height=0.33\textwidth]
{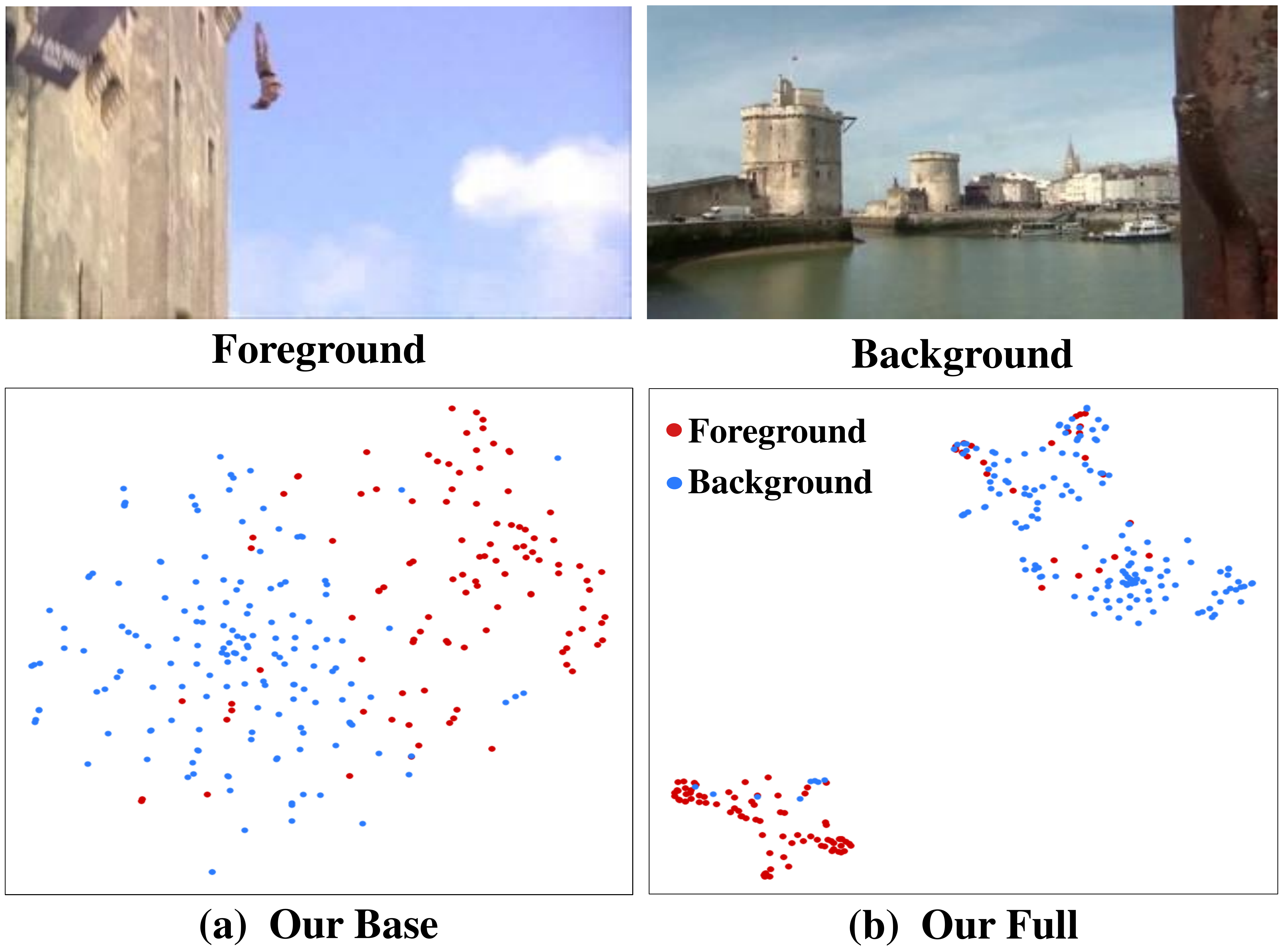}}
    \end{center}
    \vspace{-2.5mm}
    \caption{T-SNE visualization of foreground and background features on example “CliffDiving" on THUMOS14.}
\label{fig:4}
\end{figure}

\subsection{Qualitative results}
To help understand the effect of our proposed method, we present some qualitative results in this subsection. First, we show one case selected from THUMOS14 dataset in Figure~\ref{fig:3}, and we observe that our method can locate more accurate action and background regions than our Base and ASM-Loc~(black dashed boxes).
Meanwhile, we adopt t-SNE technology to project the embedding features of one video in THUMOS14 dataset into a 2-dimensional features space, and the results are shown in Figure~\ref{fig:4}. 
We observe that our method can accurately bring the embedding features of foregrounds together, and make them away from the background.
The visualization results validate the discriminative capability of the learned features and thus support the accurate estimated action localization results.

\section{Conclusion}

In this paper, we propose a novel weakly-supervised TAL method by inferring salient snippet-feature, of which several modules are designed to assist pseudo label generation by exploring the information variation and interaction. Comprehensive experiments demonstrate the effectiveness and superiority of our proposed method.

\section{Acknowledgement}
This work was partly supported by the Funds for Innovation Research Group Project of NSFC under Grant 61921003, NSFC Project under Grant 62202063, and 111 Project under Grant B18008.

\bibliography{aaai24}

\end{document}